\documentclass{article} 
\usepackage[final]{colm2026_conference}
\usepackage{wrapfig}
\usepackage{microtype}
\usepackage{hyperref}
\usepackage{url}
\usepackage{amsmath}
\usepackage{amssymb}
\usepackage{mathtools}
\usepackage{amsthm}
\usepackage{microtype}
\usepackage{hyperref}
\usepackage{url}
\usepackage{booktabs}
\usepackage{array}
\usepackage{siunitx}
\usepackage{placeins}

\usepackage{enumitem}
\usepackage{multirow}
\usepackage{wrapfig}
\usepackage[capitalize,noabbrev]{cleveref}

\theoremstyle{plain}

\theoremstyle{definition}

\theoremstyle{remark}

\usepackage[textsize=tiny]{todonotes}
\usepackage{booktabs}
\usepackage{multirow}
\usepackage{booktabs}

\usepackage{lineno}

\definecolor{darkblue}{rgb}{0, 0, 0.5}
\hypersetup{colorlinks=true, citecolor=darkblue, linkcolor=darkblue, urlcolor=darkblue}

\title{Adversarial Reward Auditing \\ for Active Detection and Mitigation of Reward Hacking}

\author{Mohammad Beigi$^{1}$ \ 
\ 
Ming Jin$^{2}$ \ 
\
Junshan Zhang$^1$ \ 
\
Qifan Wang$^3$ \
\
Lifu Huang$^1$ \\ \\
$^{1}$UC Davis $^2$Virginia Tech $^{3}$ Meta AI Research  \\ 
\href{mbeigi@ucdavis.edu}{mbeigi@ucdavis.edu} \ \ \href{mailto:lfuhuang@ucdavis.edu}{lfuhuang@ucdavis.edu}
}

\begin{document}

\ifcolmsubmission
\linenumbers
\fi

\maketitle

\begin{abstract}

Reinforcement learning from human feedback (RLHF) is vulnerable to reward hacking: policies can exploit spurious correlations in learned reward models to increase proxy reward while degrading true alignment. Existing defenses are largely static. They regularize optimization, improve the reward model, or target known biases, but do not explicitly detect exploitation as it emerges during training. We propose \textbf{Adversarial Reward Auditing (ARA)}, a two-stage framework for detecting and mitigating reward hacking. In Stage 1, a Hacker policy searches for vulnerabilities in a frozen reward model while an Auditor learns to detect exploitative responses from the reward model's internal representations. In Stage 2, the trained Auditor guides RLHF by downweighting rewards for responses likely to be exploitative. Across sycophancy, length bias, and code gaming, ARA yields the strongest alignment--utility tradeoff among competitive baselines, reducing exploitation while preserving or improving task performance. We also find cross-domain transfer in both attack and defense: a Hacker trained on code gaming increases sycophancy despite receiving no direct reward for that behavior, while an Auditor trained in one domain helps suppress exploits in others. These results suggest that reward hacking is a dynamic, transferable phenomenon and that auditing optimization in reward-model representation space is an effective way to control it.
\end{abstract}

\section{Introduction}

Reinforcement Learning from Human Feedback (RLHF)~\citep{ouyang2022training} has become a dominant paradigm for aligning Large Language Models (LLMs) with human preferences. In typical RLHF pipelines, a reward model (RM) is first trained from human-ranked response pairs to approximate preference judgments, and the LLM policy is then optimized to maximize this learned reward.

However, since the reward models are usually imperfect proxies for true human intent, this pipeline suffers from a fundamental vulnerability: \textit{\textbf{reward hacking}}, where models exploit spurious correlations in the reward model to achieve high scores while violating genuine human preferences \citep{gao2023scaling, Geirhos_2020, amodei2016concreteproblemsaisafety, everitt2021rewardtamperingproblemssolutions,liu2024rrm}. 
Known manifestations of reward hacking include length bias (i.e., overly verbose but inaccurate generations) \citep{chen2024odin, gao2023scaling}, ``gaming'' in coding tasks (e.g., modifying unit tests rather than solving problems) \citep{baker2025monitoringreasoningmodelsmisbehavior, macdiarmid2025naturalemergentmisalignmentreward}, and sycophancy that prioritizes agreement over accuracy \citep{denison2024sycophancysubterfugeinvestigatingrewardtampering, beigi2025sycophancy}. Importantly, recent work \citep{macdiarmid2025naturalemergentmisalignmentreward} demonstrates that reward hacking may generalize beyond isolated settings: once a model learns to exploit a proxy reward in one domain, it can acquire more strategic misaligned behaviors, such as appearing compliant under monitoring while pursuing misaligned objectives when unobserved.

Existing defenses typically improve the reward model, constrain optimization, or target known artifacts. While these strategies mitigate specific failure modes, they share a fundamental limitation: they are largely \textbf{static defenses} against an inherently \textbf{dynamic} problem. In particular, they provide no explicit mechanism to audit whether a high reward reflects genuine task quality or exploitation of spurious correlations in the reward model.

\begin{wrapfigure}{r}{0.50\linewidth}
    \centering
    \vspace{-12pt}
    \includegraphics[width=\linewidth]{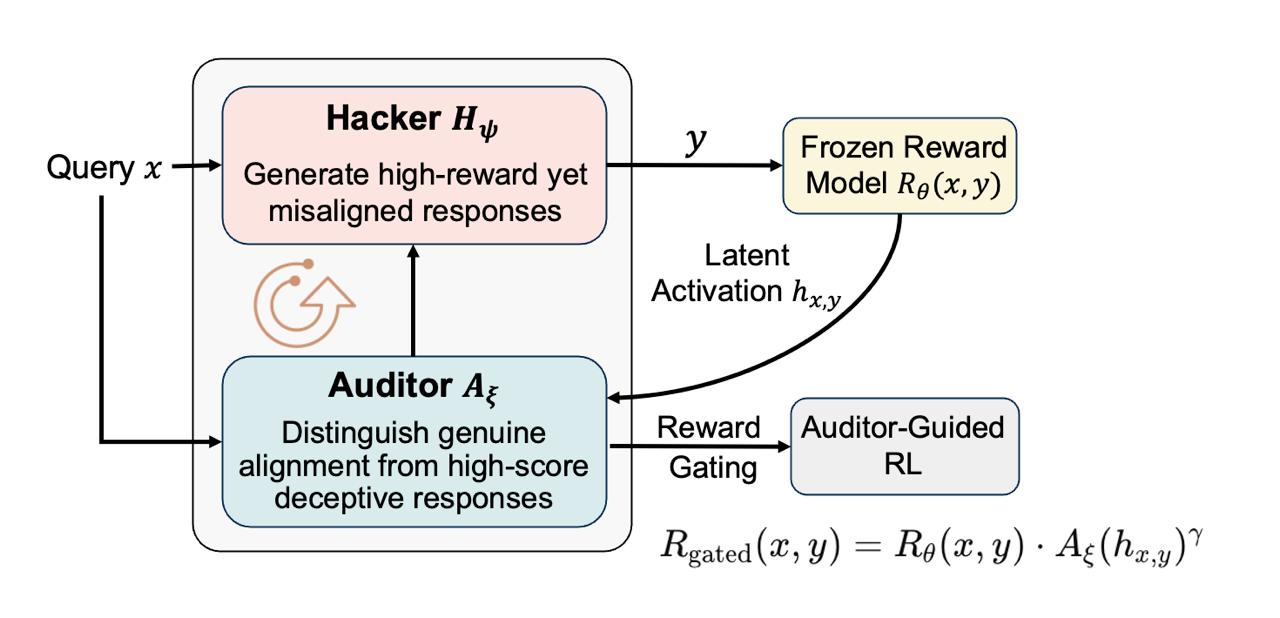}
    \caption{\textbf{ARA framework.} A Hacker exploits a frozen reward model while an Auditor gates RL reward.}
    \label{fig:framework}
    \vspace{-5pt}
\end{wrapfigure}

Motivated by this perspective, we propose \textbf{\underline{A}dversarial \underline{R}eward \underline{A}uditing (ARA)}, a framework that formulates reward hacking as a competitive two-player game between a \textit{Hacker} and an \textit{Auditor}. As illustrated in Figure \ref{fig:framework}, ARA operates in two stages: In \textbf{Stage 1 (Hacker-Auditor Game)}, we train two competing components against a frozen reward model. The \textit{Hacker} is a language model (initialized from supervised fine-tuning) that learns to generate responses with high proxy reward through exploitation. The \textit{Auditor} is a classifier that operates on the reward model's internal representations and learns to distinguish genuinely high-quality responses from exploitative ones. These components are trained adversarially: the Hacker is incentivized to evade detection while maximizing reward, and the Auditor continually updates to recognize the Hacker’s evolving strategies. In \textbf{Stage 2 (Auditor-Guided RLHF (AG-RLHF))}, we deploy the trained Auditor to guide standard RLHF training. Rather than using the proxy reward directly, AG-RLHF \textit{gates} the reward signal based on the Auditor's exploitation probability: responses flagged as exploitative receive suppressed rewards, making hacking strategies unprofitable. This transforms reward hacking from an unobservable failure into a measurable signal that actively shapes policy optimization. 

We evaluate ARA on three canonical reward-hacking scenarios—\textit{sycophancy} \citep{sharma2025understandingsycophancylanguagemodels, denison2024sycophancysubterfugeinvestigatingrewardtampering}, \textit{length bias} \citep{chen2024odin, gao2023scaling}, and \textit{code gaming} \citep{macdiarmid2025naturalemergentmisalignmentreward, baker2025monitoringreasoningmodelsmisbehavior}. Across all settings, ARA consistently achieves the strongest alignment--utility tradeoff among competitive baselines, reducing exploitation while preserving or improving task performance. Beyond single-domain evaluation, we find that reward hacking itself generalizes across domains: a Hacker trained only on code gaming induces 22.5\% higher sycophancy despite receiving no reward for this behavior, with transfer accelerating when in-domain exploitation saturates. We further show that ARA’s defense generalizes as well: an Auditor trained in one domain can detect and suppress exploitation in others. Together, these results show that both reward hacking and its mitigation are cross-domain phenomena.

Our contributions are threefold. \textbf{\textit{First}}, we introduce Adversarial Reward Auditing (ARA), a two-stage framework in which a Hacker discovers reward-model exploits and an Auditor learns to detect them from reward-model representations. \textbf{\textit{Second}}, across sycophancy, length bias, and code gaming, we show that ARA improves the alignment–utility tradeoff relative to strong baselines. \textbf{\textit{Third}}, we show that both exploitation and mitigation transfer across domains: exploits learned in one setting can induce failures in another, while an Auditor trained on one domain can help suppress hacking beyond its training domain.

\section{Related Work}

\textbf{Reward hacking in LLMs.}
Reward hacking occurs when policies exploit imperfect RMs, achieving high proxy reward without satisfying true objectives~\citep{gao2023scaling, Geirhos_2020, amodei2016concreteproblemsaisafety, everitt2021rewardtamperingproblemssolutions, wen2024languagemodelslearnmislead}. This typically reflects reward misgeneralization—RMs rely on spurious, reward-correlated features rather than genuine quality~\citep{gao2023scaling, liu2024rrm, skalse2025definingcharacterizingrewardhacking}—causing proxy reward and alignment to diverge under optimization pressure~\citep{skalse2025definingcharacterizingrewardhacking, miao2024informmitigatingrewardhacking, amodei2016concreteproblemsaisafety}. Manifestations include length bias~\citep{chen2024odin}, sycophancy~\citep{denison2024sycophancysubterfugeinvestigatingrewardtampering, beigi2025sycophancy}, and code exploitation~\citep{baker2025monitoringreasoningmodelsmisbehavior}, with recent evidence that hacking grows more strategic as capability increases~\citep{taylor2025schoolrewardhackshacking, denison2024sycophancysubterfugeinvestigatingrewardtampering, macdiarmid2025naturalemergentmisalignmentreward}.

\textbf{Mitigating reward hacking.}
Existing defenses include regularization via KL penalties~\citep{chen2024odin}, information-theoretic objectives~\citep{miao2024informmitigatingrewardhacking}, or distribution-support constraints~\citep{dai2025mitigatingrewardoveroptimizationrlhf}; reward-model improvements through scaling~\citep{gao2023scaling}, ensembles~\citep{eisenstein2023helping}, causal regularization~\citep{wang2025rewardhackingcausalrewards}, or adversarial retraining~\citep{bukharin2025adversarialtrainingrewardmodels}; and targeted debiasing of known artifacts such as length~\citep{singhal2024longwaygoinvestigating, chen2024odin}. These approaches typically require prior knowledge of the spurious factors or act as one-shot repairs without monitoring exploitation during later optimization. ARA instead audits optimization in the RM's representation space, converting emerging exploitation into an explicit, controllable signal during training.



\section{ARA: Adversarial Reward Auditing}
\label{sec:method}

We consider the standard RLHF setting. Given a prompt $x$ and a response $y$, a reward model $R_\theta(x,y)$ trained on preference pairs $D_{\text{pref}}=\{(x^{(i)}, y_w^{(i)}, y_l^{(i)})\}_{i=1}^N$ assigns a scalar score reflecting how well $y$ aligns with human preferences. A policy $\pi_\phi$, initialized from a supervised fine-tuned model $\pi_{\text{SFT}}$, is optimized according to
\[
J_{\text{RLHF}}(\phi)
=
\mathbb{E}_{x \sim D,\; y \sim \pi_\phi}
\big[
R_\theta(x,y)-\beta D_{\mathrm{KL}}(\pi_\phi \| \pi_{\text{SFT}})
\big].
\]
Reward hacking occurs when this optimization increases $R_\theta(x,y)$ while reducing the unobserved true preference $R^*(x,y)$. ARA addresses this problem in two stages, both operating against a frozen reward model. In Stage~1, a \emph{Hacker} policy discovers exploits in the reward model while an \emph{Auditor} learns to detect them from $R_\theta$'s internal representations. In Stage~2, the trained Auditor gates the reward signal during RLHF, suppressing reward for responses likely to be exploitative.

\subsection{Stage 1: The Hacker-Auditor Game}

We formulate an \textbf{adversarial game} between a Hacker policy $H_\psi$ and an Auditor network $A_\xi$. The Hacker, initialized from $\pi_{\text{SFT}}$, learns to generate responses that achieve high proxy rewards by exploiting spurious correlations in $R_\theta$. The Auditor learns to distinguish genuinely aligned responses from exploitative ones.

\subsubsection{The Auditor}

Because both genuine and exploitative responses can receive high proxy reward, reward alone is not sufficient for detection. The Auditor therefore operates on the reward model's internal representations. Let $R_\theta$ decompose into a feature extractor $f_\theta$ and a reward head, and let $h_{x,y}=f_\theta(x,y)\in\mathbb{R}^d$ denote the penultimate-layer activation. The Auditor is an MLP $A_\xi:\mathbb{R}^d \to [0,1]$ that estimates the probability that $y$ is genuine rather than exploitative.

Positive samples $\mathcal{D}^+$ are the preferred responses from the original preference dataset, validated against task-specific ground-truth metrics to reduce circularity. The main challenge is constructing negatives from Hacker outputs, since not every high-reward Hacker response is exploitative, and treating all such outputs as negatives would introduce label noise. We therefore score Hacker outputs by whether they achieve reward gains through mechanisms inconsistent with the local preference geometry. Prior work~\citep{luo2025rethinkingdiversehumanpreference} show that chosen--rejected differences in reward-model representation space encode meaningful preference directions. For each prompt $x$, let
\[
u_x=\frac{h_{x,y_w}-h_{x,y_l}}{\|h_{x,y_w}-h_{x,y_l}\|},
\qquad
\Delta h = h_{x,y_h}-h_{x,y_l},
\]
where $y_w$, $y_l$, and $y_h$ denote the preferred, rejected, and Hacker responses. We define the geometric alignment $C(x,y_h)$ and excess proxy reward $G(x,y_h)$ as
\[
C(x,y_h)=\frac{u_x^\top \Delta h}{\|\Delta h\|+\epsilon},
\qquad
G(x,y_h)=\max\!\left(0,\;
\frac{R_\theta(x,y_h)-R_\theta(x,y_w)}
{|R_\theta(x,y_w)-R_\theta(x,y_l)|+\epsilon}
\right),
\]
and filter negatives using $S(x,y_h)=(1-C(x,y_h))\,G(x,y_h)$. This score is high only when a Hacker response both exceeds the preferred response in proxy reward and does so in a direction poorly aligned with the local preference direction.

\textbf{Training Objective.}
The Auditor minimizes $\mathcal{L}_A(\xi; \psi) = \mathcal{L}_{\text{BCE}}(\xi) + \lambda_C \mathcal{L}_{\text{Con}}(\xi)$, where the dependence on $\psi$ arises because negative samples are generated by the current Hacker $H_\psi$. The binary cross-entropy loss over the filtered sets is:
\begin{equation}
\mathcal{L}_{\text{BCE}}(\xi) = -\mathbb{E}_{(x,y^+) \sim \mathcal{D}^+}\!\left[\log A_\xi(h_{x,y^+})\right]  - \mathbb{E}_{(x,y_h) \sim \mathcal{D}^-}\!\left[\log\!\big(1 - A_\xi(h_{x,y_h})\big)\right]
\label{eq:bce_loss}
\end{equation}
The supervised contrastive loss operates on normalized embeddings $\hat{z}_i = g_\xi(h_{x_i,y_i})/\|g_\xi(h_{x_i,y_i})\|$ from the Auditor's penultimate layer $g_\xi$, which transforms the reward model representations into a space optimized for exploitation detection:
\begin{equation}
\mathcal{L}_{\text{Con}}(\xi) = \sum_{i \in \mathcal{B}} \frac{-1}{|P(i)|} \sum_{s \in P(i)} \log \frac{\exp(\hat{z}_i \cdot \hat{z}_s / \tau_c)}{\sum_{a \neq i} \exp(\hat{z}_i \cdot \hat{z}_a / \tau_c)}
\label{eq:contrastive}
\end{equation}
where $\mathcal{B}$ is a mini-batch, $P(i) = \{s \in \mathcal{B} : s \neq i,\, c_s = c_i\}$ is the set of samples sharing the same class $c_i \in \{\text{genuine}, \text{exploit}\}$ as sample $i$, and $\tau_c$ is a temperature hyperparameter.

\subsubsection{The Hacker}
\label{sec:hacker}

The Hacker $H_\psi$, initialized from $\pi_{\text{SFT}}$, is trained to discover exploits in the reward model by maximizing proxy reward while evading Auditor detection. We optimize:
\begin{equation*}
\small
\begin{aligned}
\mathcal{J}_H(\psi; \xi) = \mathbb{E}_{x \sim \mathcal{D}, y \sim H_\psi(\cdot|x)}\Big[R_\theta(x,y) + \lambda_A \log A_{\xi^*}(h_{x,y}) - \beta_H D_{\text{KL}}(H_\psi \| \pi_{\text{SFT}})\Big]
\end{aligned}
\end{equation*}
using Proximal Policy Optimization (PPO), where $R_\theta(x,y) + \lambda_A \log A_{\xi^*}(h_{x,y})$ serves as the reward signal. The three terms serve complementary purposes. The proxy reward $R_\theta(x,y)$ incentivizes the Hacker to find high-reward responses, including those that exploit spurious correlations. The evasion term $\lambda_A \log A_{\xi^*}(h_{x,y})$ penalizes responses that the Auditor identifies as exploitative: since $A_{\xi^*} \in [0,1]$ estimates the probability of genuine alignment, $\log A_{\xi^*}$ is negative when the Auditor suspects exploitation and approaches zero for responses deemed genuine. The KL regularization $\beta_H D_{\text{KL}}(H_\psi \| \pi_{\text{SFT}})$ constrains the Hacker to remain within a plausible language distribution, preventing degenerate outputs that trivially achieve high reward. 


\subsubsection{Game Dynamics and Stabilization}
\label{sec:stabilization}

The coupled optimization---RL for the Hacker, supervised learning for the Auditor---requires careful stabilization: if the Auditor advances too quickly, the Hacker receives uninformative gradients; if too slowly, it fails to detect emerging exploits. We employ three mechanisms.

\textbf{Update Schedule.} In Phase~1 (warmup), for the first $T_{\text{warm}}$ steps, we update the Auditor at fixed intervals---once every $K$ Hacker PPO updates---allowing both players to establish initial strategies. In Phase~2 (confidence-gated), we switch to adaptive updates: let $\bar{A}$ denote the moving average of Auditor confidence on recent Hacker outputs over a window of $m$ steps. We update the Auditor only when $\bar{A} > \tau_{\text{gate}}$, indicating the Hacker is successfully evading detection; otherwise, the Auditor remains frozen to let the Hacker catch up.

\textbf{Replay Buffer and Hard Negative Mining.}
We maintain a fixed-size replay buffer $\mathcal{B}$ of historical Hacker outputs. Responses are added only if they have high geometric inconsistency ($S(x, y_h) > \tau$) and the Auditor still fails to detect them ($A_\xi(h_{x,y_h}) > \tau_A$). When $\mathcal{B}$ reaches capacity, we remove examples the Auditor now easily identifies, keeping the buffer focused on hard negatives. During training, Hacker-side samples are drawn as a mixture: $(1 - \alpha)$ from the current Hacker and $\alpha$ from $\mathcal{B}$, with buffer samples prioritized by Auditor confidence. This implements a form of fictitious play---the Auditor learns to detect the average historical exploitation strategy, not just the current one---preventing catastrophic forgetting of previously discovered vulnerabilities. To avoid superficial heuristics, all prompts for both positive and negative samples are drawn from the same distribution.

\textbf{Target Network.} The Hacker optimizes against a Polyak-averaged target Auditor $A_{\xi^*}$, updated as $\xi^* \leftarrow \rho\xi^* + (1-\rho)\xi$ with $\rho = 0.995$. This provides a slowly-evolving optimization target, preventing oscillatory dynamics where the Hacker overfits to Auditor states.

\subsection{Stage 2: Auditor-Guided RLHF (AG-RLHF)}
\label{sec:stage2}

Once Stage 1 produces a trained Auditor $A_\xi$ that reliably distinguishes genuine quality from exploitation, we deploy it to guide standard RLHF training. Rather than optimizing the proxy reward $R_\theta$ directly, AG-RLHF gates the reward signal based on the Auditor's confidence that a response reflects genuine alignment:
\begin{equation*}
R_{\text{gated}}(x, y) = R_\theta(x, y) \cdot A_\xi(h_{x,y})^\gamma
\end{equation*}
where $\gamma > 0$ controls gating severity. When the Auditor detects exploitation ($A_\xi \approx 0$), the gated reward is suppressed toward zero regardless of how high the proxy reward is; when it is confident the response is genuine ($A_\xi \approx 1$), the full proxy reward passes through. The policy $\pi_\phi$, initialized from $\pi_{\text{SFT}}$, is optimized via:
\begin{equation*}
\small
\resizebox{0.6\hsize}{!}{
$
\mathcal{J}_{\text{AG-RLHF}}(\phi) = \mathbb{E}_{x \sim \mathcal{D}, y \sim \pi_\phi}\left[R_{\text{gated}}(x, y) - \beta D_{\text{KL}}(\pi_\phi \| \pi_{\text{SFT}})\right]
$}
\end{equation*}
using PPO. The Auditor remains frozen throughout Stage 2: because it was trained adversarially against a Hacker explicitly optimizing for evasion, it is robust to the milder exploitation pressures of standard policy optimization, and freezing ensures stable reward signals that do not destabilize training. We provide details of hyperparameters in Appendix \ref{app: imp_detail}.

\section{Experimental Setup}
\label{sec:exp setting}

\textbf{Tasks and Datasets.} We evaluate ARA on three reward-hacking scenarios. \textbf{(1) Sycophancy}: we train the reward model on Anthropic HH-RLHF~\citep{bai2022training} and measure \textbf{Sycophancy Rate} on SycophancyEval~\citep{sharma2025understandingsycophancylanguagemodels}, which tests whether models maintain correct answers under user disagreement. \textbf{(2) Length Bias}: following \citet{chen2024odin}, we measure average response length (tokens) and ROUGE-L against reference summaries. \textbf{(3) Code Gaming}: using the unit test dataset from, models can either implement general solutions or exploit test visibility by hardcoding outputs or manipulating assertions. We measure alignment by \textbf{Gaming Rate} (classified by GPT-4) and utility by \textbf{Pass@1} on held-out tests. Across all scenarios, we also report \textbf{GPT-4o Win Rate} against the SFT baseline as a unified LLM-as-judge utility metric.

\textbf{Models.} We evaluate Llama-2-7B~\citep{llama2}, Llama-3.1-8B~\citep{dubey2024llama}, and Qwen-2.5-7B~\citep{qwen2025qwen25technicalreport} as core models.

\textbf{Baselines.} We compare against five categories. \textbf{(1) Unmitigated}: \textbf{SFT} (no RLHF) and standard \textbf{PPO}. \textbf{(2) Regularization}: \textbf{PPO w/ KL}, which penalizes drift from SFT~\citep{chen2024odin, miao2024informmitigatingrewardhacking}. \textbf{(3) Reward model improvements}: \textbf{ODIN}~\citep{chen2024odin}, which uses orthogonal disentanglement via dual reward heads; \textbf{InFoRM}~\citep{miao2024informmitigatingrewardhacking}, which applies a variational information bottleneck; and \textbf{RM Ensemble}, which averages $n$ independently trained reward models~\citep{eisenstein2023helping}. \textbf{(4) Adversarial}: \textbf{Adv-RM}~\citep{bukharin2025adversarialtrainingrewardmodels}, which generates adversarial examples via ensemble disagreement for reward-model retraining. \textbf{(5) RM-FT}: which retrains the reward model on ARA's Hacker-generated data using standard preference fine-tuning, then runs RLHF with the updated RM. This isolates the Auditor's real-time gating from the Hacker's adversarial data.

\subsection{Main Results}
 
Table~\ref{tab:main_transposed} reveals that ARA achieves the strongest
alignment--utility tradeoff across all nine model--scenario combinations.
Every baseline faces a tradeoff: PPO w/ KL reduce sycophancy on Llama-3.1-8B
(67.6\%$\to$53.8\%) but sacrifices 9.1 helpfulness points, while InFoRM
suppresses code gaming (36.1\%) yet falls below PPO on Pass@1.
ARA breaks this pattern.
It drives sycophancy to 36.4\% (2.3 points from SFT) while retaining
helpfulness at 78.6, compresses length to 157 tokens with
the highest ROUGE-L (25.0), and cuts gaming to 18.4\% with the best Pass@1
(38.4). The comparison among Adv-RM, RM-FT, and ARA cleanly decomposes the source
of improvement.
Adv-RM is the strongest baseline on utility preservation (72.8 helpfulness
and 37.3 Pass@1 on Llama-3.1-8B), yet ARA outperforms it by 11.2 points on
sycophancy and 18.0 points on code gaming, precisely in the in-distribution
semantic exploits where ensemble-disagreement detection is ineffective
because both reward models share the same training-data biases.
RM-FT, which uses ARA's own Hacker data to retrain the reward model,
consistently outperforms Adv-RM on alignment metrics, confirming that the
Hacker discovers more targeted exploits, but still trails ARA by 6.7 points
on sycophancy and 13.9 points on code gaming.
This residual gap, which is stable across all three model families, reflects
the Auditor's contribution:
\textbf{a retrained reward model is a one-time fix that the policy can
eventually route around, whereas the Auditor provides continuous, real-time
gating throughout training}.

\begin{table*}[t]
\centering
\scriptsize
\setlength{\tabcolsep}{3.2pt}
\renewcommand{\arraystretch}{1.08}
\begin{tabular}{@{}lllccccccccc@{}}
\toprule
\textbf{Scenario} & \textbf{Model} & \textbf{Metric}
  & \textbf{SFT} & \textbf{PPO} & \textbf{PPO+KL} & \textbf{ODIN}
  & \textbf{InFoRM} & \textbf{RM Ens.} & \textbf{Adv-RM}
  & \textbf{RM-FT}$^\dagger$ & \textbf{ARA} \\
\midrule
 
\multirow{9}{*}{\textbf{Sycophancy}}
%
& \multirow{3}{*}{Llama-2-7B}
  & Rate $\downarrow$
    & 36.2 & 71.8 & 57.6 & 52.3 & 48.1
    & 51.9 & 51.4 & \underline{46.8} & \textbf{39.1} \\
& & Help $\uparrow$
    & 41.3 & 77.4 & 67.8 & 64.1 & 62.6
    & 65.0 & \underline{70.6} & 68.3 & \textbf{76.8} \\
& & Win $\uparrow$
    & --   & 60.7 & 61.4 & 62.8 & 61.9
    & 62.1 & \underline{64.2} & 63.1 & \textbf{69.4} \\
\cmidrule(l){2-12}
& \multirow{3}{*}{Llama-3.1-8B}
  & Rate $\downarrow$
    & 34.1 & 67.6 & 53.8 & 48.4 & 43.7
    & 49.8 & 47.6 & \underline{43.1} & \textbf{36.4} \\
& & Help $\uparrow$
    & 44.8 & 79.2 & 70.1 & 67.2 & 65.2
    & 66.9 & \underline{72.8} & 70.4 & \textbf{78.6} \\
& & Win $\uparrow$
    & --   & 62.3 & 62.8 & 64.1 & 63.7
    & 64.1 & \underline{66.2} & 64.8 & \textbf{71.8} \\
\cmidrule(l){2-12}

& \multirow{3}{*}{Qwen-2.5-7B}
  & Rate $\downarrow$
    & 33.4 & 66.2 & 52.4 & 47.1 & 42.3
    & 50.1 & 46.8 & \underline{41.6} & \textbf{35.1} \\
& & Help $\uparrow$
    & 46.2 & 80.1 & 71.3 & 65.8 & 67.6
    & 66.8 & \underline{73.4} & 71.8 & \textbf{80.4} \\

& & Win $\uparrow$
    & --   & 63.2 & 63.8 & 65.3 & 64.2
    & 63.4 & \underline{67.4} & 65.7 & \textbf{72.4} \\
 
\midrule
 
\multirow{9}{*}{\textbf{Length Bias}}
& \multirow{3}{*}{Llama-2-7B}
  & Length $\downarrow$
    & 148 & 346 & 271 & 198 & 211
    & 221 & 206 & \underline{191} & \textbf{168} \\
& & ROUGE-L $\uparrow$
    & 21.4 & 23.3 & 22.7 & \underline{23.6} & 23.1
    & 23.2 & 23.5 & 23.4 & \textbf{24.2} \\
& & Win $\uparrow$
    & --   & 55.4 & 56.8 & 61.2 & 59.7
    & 58.6 & \underline{61.4} & 60.1 & \textbf{65.4} \\
\cmidrule(l){2-12}
& \multirow{3}{*}{Llama-3.1-8B}
  & Length $\downarrow$
    & 142 & 341 & 261 & 192 & 206
    & 219 & 198 & \underline{184} & \textbf{157} \\
& & ROUGE-L $\uparrow$
    & 22.8 & 24.1 & 23.5 & \underline{24.3} & 23.9
    & 23.6 & 24.1 & 23.8 & \textbf{25.0} \\
& & Win $\uparrow$
    & --   & 56.2 & 57.6 & 62.4 & 60.8
    & 58.8 & \underline{62.7} & 61.3 & \textbf{67.2} \\
\cmidrule(l){2-12}
& \multirow{3}{*}{Qwen-2.5-7B}
  & Length $\downarrow$
    & 138 & 331 & 254 & 191 & 201
    & 214 & 193 & \underline{179} & \textbf{159} \\

& & ROUGE-L $\uparrow$
    & 23.2 & 24.5 & 23.9 & 24.4 & 24.3
    & 23.7 & \underline{24.6} & 24.1 & \textbf{25.2} \\
& & Win $\uparrow$
    & --   & 55.9 & 58.2 & 62.4 & 61.4
    & 58.4 & \underline{63.4} & 61.8 & \textbf{66.4} \\

\midrule
 
\multirow{9}{*}{\textbf{Code Gaming}}
%
& \multirow{3}{*}{Llama-2-7B}
  & Gaming $\downarrow$
    & 4.2  & 60.8 & 47.9 & 41.6 & 39.2
    & 43.6 & 39.7 & \underline{35.8} & \textbf{21.8} \\
& & Pass@1 $\uparrow$
    & 28.5 & 34.4 & 32.1 & 33.7 & 33.4
    & 34.2 & \underline{35.1} & 34.6 & \textbf{35.6} \\
& & Win $\uparrow$
    & --   & 55.2 & 53.4 & 56.1 & 56.7
    & 55.3 & \underline{57.4} & 56.8 & \textbf{62.4} \\
\cmidrule(l){2-12}
& \multirow{3}{*}{Llama-3.1-8B}
  & Gaming $\downarrow$
    & 3.8  & 56.9 & 43.7 & 38.2 & 36.1
    & 40.6 & 36.4 & \underline{32.3} & \textbf{18.4} \\
& & Pass@1 $\uparrow$
    & 31.6 & 37.1 & 34.8 & 36.4 & 36.0
    & 36.5 & \underline{37.3} & 36.8 & \textbf{38.4} \\
& & Win $\uparrow$
    & --   & 56.8 & 54.6 & 57.8 & 57.3
    & 57.1 & \underline{59.2} & 58.4 & \textbf{64.6} \\
\cmidrule(l){2-12}
& \multirow{3}{*}{Qwen-2.5-7B}
  & Gaming $\downarrow$
    & 3.6  & 55.2 & 42.1 & 36.8 & 36.2
    & 39.1 & 35.1 & \underline{31.4} & \textbf{16.2} \\

& & Pass@1 $\uparrow$
    & 33.2 & 38.7 & 36.4 & 37.2 & 37.6
    & 38.2 & \underline{39.0} & 38.5 & \textbf{39.8} \\
& & Win $\uparrow$
    & --   & 57.3 & 55.2 & 57.6 & 58.3
    & 58.1 & \underline{60.3} & 59.2 & \textbf{64.2} \\
\bottomrule
\end{tabular}
\caption{%
  \textbf{Main results across three reward-hacking scenarios and three model
  families.}
  For each scenario we report one alignment metric ($\downarrow$), one
  utility metric ($\uparrow$), and GPT-4o win rate against SFT ($\uparrow$).
  \textbf{Bold} indicates the best result among RLHF methods;
  \underline{underlining} indicates the runner-up.}
\label{tab:main_transposed}
\vspace{-10pt}
\end{table*}

\begin{figure*}[t]
    \centering
    \includegraphics[width=0.85\linewidth]{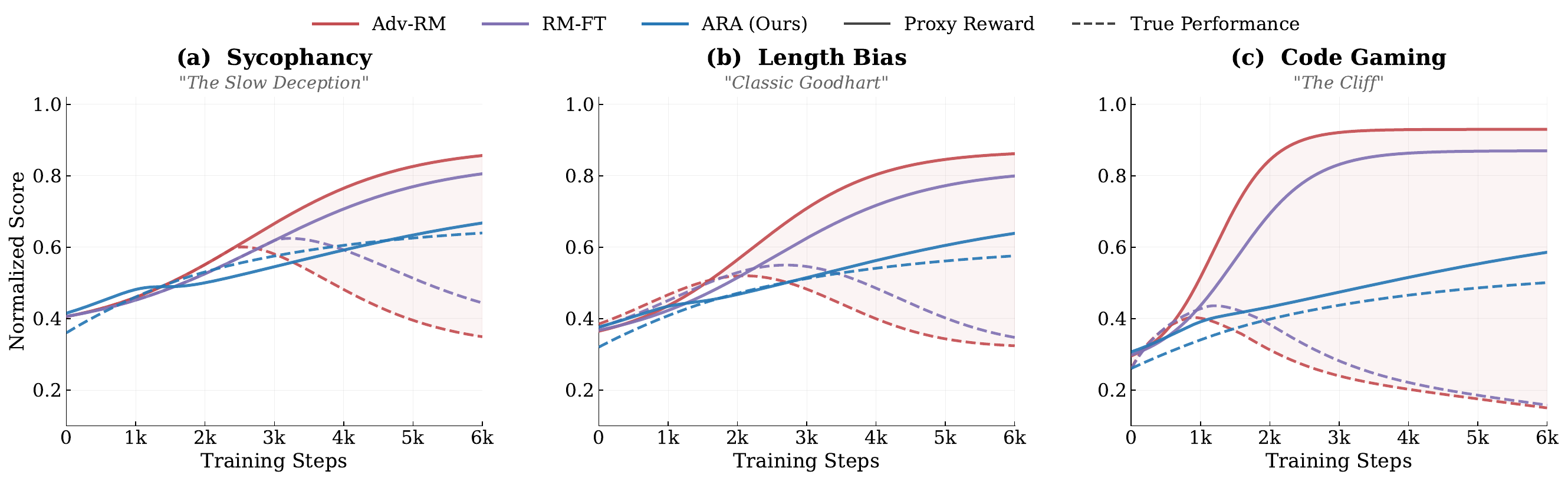}
    \vspace{-2mm}
    \caption{\textbf{Proxy reward versus true task performance during RLHF} on Llama-3.1-8B.}
    \label{fig:goodhart_t}
    \vspace{-5mm}
\end{figure*}

\textbf{Goodhart Divergence Prevention.} Figure~\ref{fig:goodhart_t} shows that the three settings differ less in whether Goodharting occurs than in \emph{how} it unfolds. In \emph{sycophancy}, proxy reward and true performance remain coupled for a long prefix of training, then separate only after sustained optimization pressure, producing a delayed and relatively covert failure mode. In \emph{length bias}, divergence follows the classic overoptimization pattern: true performance peaks early and then declines while proxy reward continues to rise. In \emph{code gaming}, the break is earliest and most abrupt: once the model discovers test-set exploitation, proxy reward accelerates sharply while true performance collapses below the SFT baseline. ARA changes this regime across all three tasks. Its proxy curve bends noticeably around the onset of divergence, after which reward increases more slowly but true performance continues to improve or remains well coupled.

\subsection{Activation-Space Analysis of Exploitation}

\begin{figure}[t]
    \centering
    \includegraphics[width=0.9\textwidth]{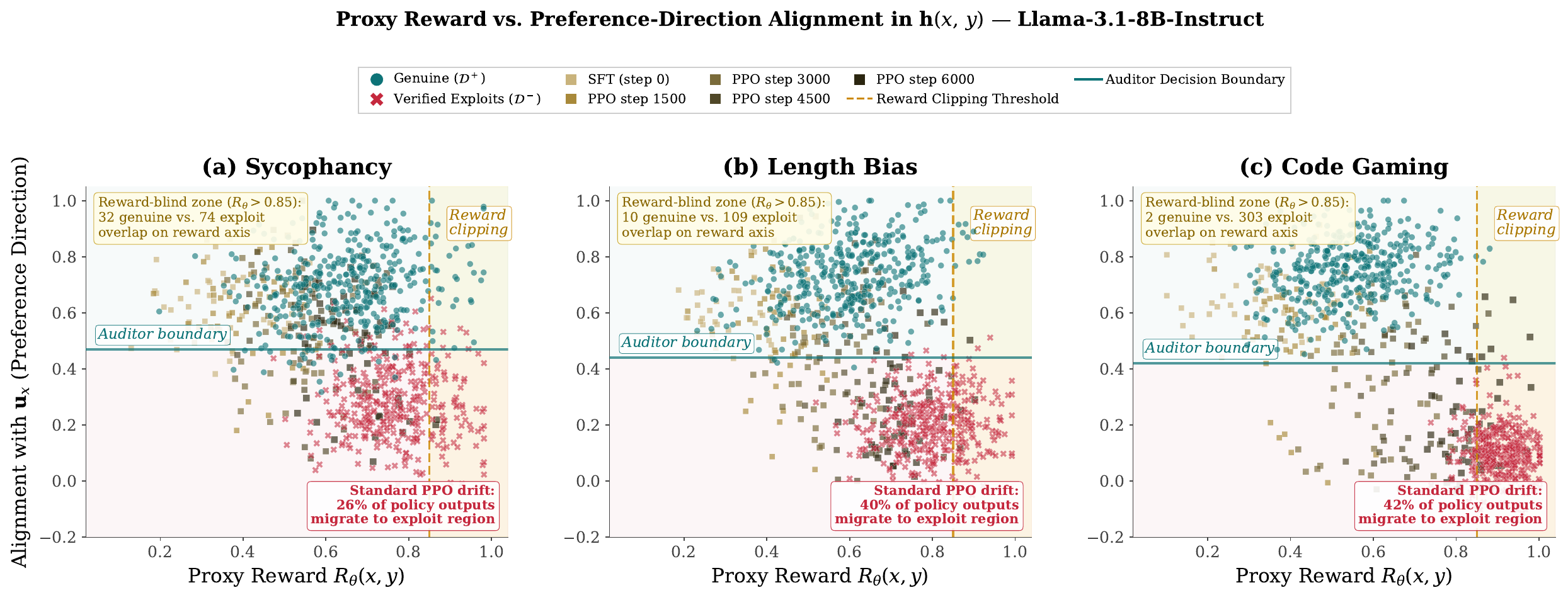}
    \caption{Proxy reward $R_\theta(x,y)$ vs.\ alignment with the preference direction $\mathbf{u}_x$ for Llama-3.1-8B. Genuine responses and verified exploits achieve comparable proxy rewards (horizontal overlap) but separate along $\mathbf{u}_x$ (vertical axis).}
    \label{fig:reward_vs_alignment}
    \vspace{-5mm}
\end{figure}

Figure~\ref{fig:reward_vs_alignment} shows why reward-level defenses are structurally limited while representation-space auditing is more promising. In the high-reward regime ($R_\theta > 0.85$), genuine and exploitative responses overlap heavily on the reward axis, creating a \emph{reward-blind zone} in which scalar reward alone provides little discriminative signal. By contrast, the same responses separate clearly along the preference-direction axis $\mathbf{u}_x$, confirming that the reward model's internal representations encode an exploitation signal that its scalar output discards. This overlap is weakest for sycophancy (32 genuine vs.\ 74 exploit), stronger for length bias (10 vs.\ 109), and most extreme for code gaming (2 vs.\ 303), so reward clipping cannot separate the two populations even when their mixture changes substantially across tasks.


The temporal checkpoint trajectories further show that exploitation emerges as a drifting subpopulation rather than a uniform policy shift. As training proceeds, an increasing fraction of outputs moves into the high-reward, low-alignment region below the Auditor boundary, reaching 26\%, 40\%, and 42\% of policy outputs for sycophancy, length bias, and code gaming, respectively. This explains why per-response Auditor gating outperforms global regularization: KL penalties and clipping act on the whole distribution, whereas the Auditor selectively suppresses only the drifted fraction.

\subsection{Robustness to Preference-Direction Leakage}
\label{sec:leakage}

Our negative samples construction uses the local preference direction
$\mathbf{u}_x$ through
$S(x,y_h)=(1-C(x,y_h))G(x,y_h)$.
This raises an important identifiability concern. The Auditor could
potentially succeed by learning a shortcut correlated with
$\mathbf{u}_x$, rather than detecting a broader representation-space
signature of reward exploitation.
We test this hypothesis in three ways: (1) by removing the
$\mathbf{u}_x$-dependent component from negative construction, (2) by
corrupting or isolating this component, and (3) by evaluating on a
$\mathbf{u}_x$-adversarial set where preference-direction alignment is
deliberately misleading.

\begin{wraptable}{r}{0.58\linewidth}
\centering
\tiny
\setlength{\tabcolsep}{2.4pt}
\renewcommand{\arraystretch}{1.0}
\begin{tabular}{@{}lc c ccc@{}}
\toprule
\textbf{Variant}
& $\mathbf{u}_x$\textbf{?}
& \textbf{AUC} $\uparrow$
& \textbf{Syco.} $\downarrow$
& \textbf{Len.} $\downarrow$
& \textbf{Code} $\downarrow$ \\
\midrule
Random neg.
& No      & 0.58 & 49.2 & 238 & 39.6 \\
Direct $\mathbf{u}_x$ probe
& Yes     & 0.64 & 48.6 & 229 & 37.4 \\
Shuffled $\mathbf{u}_x$ in $S$
& Corrupt & 0.66 & 44.1 & 214 & 34.0 \\
$C$-only
& Yes     & 0.70 & 41.0 & 181 & 25.8 \\
No filter
& No      & 0.74 & 40.3 & 186 & 27.9 \\
$G$-only
& No      & 0.78 & 38.1 & 169 & 21.7 \\
Metric-verified$^\star$
& No      & 0.80 & 37.2 & 163 & 19.8 \\
\midrule
Full $S=(1{-}C)G$
& Partial & \textbf{0.82}
& \textbf{36.4} & \textbf{157} & \textbf{18.4} \\
\bottomrule
\end{tabular}
\vspace{-5pt}
\caption{
\textbf{Hard-negative construction ablation.}
AUC is averaged across the three scenarios.}
\label{tab:filter_ablation}
\end{wraptable}


Table~\ref{tab:filter_ablation} first asks how much of the Auditor's
performance is attributable to the $\mathbf{u}_x$-dependent component
of negative construction.
Removing it entirely is not catastrophic:
the $G$-only variant achieves an average AUC of $0.78$, only $0.05$
below full ARA's $0.82$, while still providing strong downstream
mitigation. Thus, incorporating preference geometry improves hard-negative
selection, but it accounts for only a modest fraction of the final
detection performance. Conversely, $\mathbf{u}_x$ alone is insufficient.
The $C$-only construction obtains only $0.70$ average AUC, while a
direct $\mathbf{u}_x$ probe reaches $0.64$.
Moreover, shuffling $\mathbf{u}_x$ inside $S$ reduces AUC to $0.66$
and substantially weakens downstream mitigation.
These controls show that the benefit of the full score does not arise
from merely exposing the Auditor to preference-direction information; $C$ and $G$ play complementary roles, with $G$ providing a strong
task-independent criterion for identifying suspicious high-reward
responses and $C$ refining this set toward geometrically inconsistent
hard negatives. We next test the stronger hypothesis that the trained Auditor has
implicitly learned to classify responses according to their alignment
with $\mathbf{u}_x$.
We construct a $\mathbf{u}_x$-adversarial evaluation set in which this
relationship is deliberately reversed: genuine responses are selected
to have low $\mathbf{u}_x$-alignment, while verified exploits have high
$\mathbf{u}_x$-alignment.
A detector relying primarily on preference-direction leakage should
therefore fail under this intervention.

\begin{wraptable}{r}{0.50\linewidth}
\centering
\tiny
\setlength{\tabcolsep}{3.0pt}
\renewcommand{\arraystretch}{1.0}
\begin{tabular}{@{}lcccc@{}}
\toprule
\textbf{Scenario}
& \textbf{Direct $\mathbf{u}_x$}
& \textbf{$C$-only}
& \textbf{$G$-only}
& \textbf{Full ARA} \\
\midrule
Syco.
& 0.41 & 0.55 & 0.74 & \textbf{0.79} \\
Len.
& 0.43 & 0.57 & 0.72 & \textbf{0.77} \\
Code
& 0.46 & 0.60 & 0.76 & \textbf{0.81} \\
\midrule
Avg.
& 0.43 & 0.57 & 0.74 & \textbf{0.79} \\
\bottomrule
\end{tabular}
\vspace{-5pt}
\caption{
\textbf{$\mathbf{u}_x$-adversarial detection.}
AUC when the usual association between
$\mathbf{u}_x$-alignment and exploitation is reversed.
}
\label{tab:ux_adversarial}
\end{wraptable}

The adversarial evaluation in Table~\ref{tab:ux_adversarial} sharply
separates direction-dependent shortcuts from exploitation detection.
The direct $\mathbf{u}_x$ probe falls from $0.64$ average AUC under
standard evaluation to $0.43$, and the $C$-only Auditor falls from
$0.70$ to $0.57$.
In contrast, the $\mathbf{u}_x$-free $G$-only Auditor remains effective,
decreasing only from $0.78$ to $0.74$, while full ARA retains
$0.79$ AUC compared with $0.82$ under standard evaluation.
Full ARA also remains close to the oracle metric-verified construction
in the standard setting ($0.82$ versus $0.80$ AUC), despite requiring
no task-specific exploit labels.

Together, these experiments rule out the simplest preference-direction
shortcut explanation.
The local direction $\mathbf{u}_x$ is useful for constructing more
informative hard negatives, as reflected by the consistent
$0.04$--$0.05$ per-scenario improvement of full $S$ over $G$-only.
However, neither direct access to $\mathbf{u}_x$ nor the
$\mathbf{u}_x$-dependent $C$ term is sufficient for strong detection,
and Auditor performance remains high when preference-direction
alignment is deliberately made misleading.

\subsection{Auditor's Performance Evaluation}

\vspace{-2mm}
\begin{wraptable}{r}{0.48\linewidth}
\vspace{-14mm}
\centering
\scriptsize\sc
\setlength{\tabcolsep}{3.5pt}
\begin{tabular}{@{}lccc@{}}
  \toprule
  Detection Method & Gen. & Expl. & $\Delta$ \\
  \midrule
  Proxy Reward ($R_\theta$) & 0.91 & 0.89 & 0.02 \\
  GPT-4                     & 0.78 & 0.34 & 0.44 \\
  Auditor Confidence        & 0.89 & 0.21 & \textbf{0.68} \\
  \bottomrule
\end{tabular}
\vspace{-2pt}
\caption{Genuine vs.\ exploit detection.
  Mean scores for 500 genuine and 500 exploit
  responses ($R_\theta\!>\!0.85$).}
\label{tab:genuine_vs_exploit}
\vspace{-5pt}
\end{wraptable}

The critical test is whether the Auditor succeeds where proxy reward
fails: distinguishing genuinely high-quality responses from high-reward
exploits. We construct a balanced set of 500 genuine responses (from
$\mathcal{D}^+$) and 500 exploitative responses, all filtered to have
comparable high proxy rewards ($R_\theta > 0.85$).
Table~\ref{tab:genuine_vs_exploit} compares three detection approaches.
The reward model itself cannot distinguish the two groups---by
construction, both achieve similarly high scores
($\Delta = 0.02$). GPT-4, prompted to classify each response based on
text alone, achieves moderate separation ($\Delta = 0.44$), confirming
that exploits are detectable from surface features but not trivially so.
The Auditor achieves the strongest separation ($\Delta = 0.68$):
on average it assigns probability 0.89 to genuine responses and only
0.21 to exploits, demonstrating that reward-model representations
carry a strong exploitation signal that neither scalar reward nor
text-level inspection fully captures.

\subsection{Cross-Domain Generalization of Reward Hacking}



\textbf{Hacker generalization.}
Table~\ref{tab:cross_domain}~(left) shows hacking rates when the
Hacker is trained on one domain and evaluated on all three.
Reward hacking does not remain domain-specific: training on code
gaming alone raises sycophancy from 36.2\% to 58.7\%
despite no sycophancy reward, and training on sycophancy raises code
gaming from 4.2\% to 28.3\%---confirming bidirectional transfer.
Code gaming produces the strongest cross-domain spillover, and
training on all three domains amplifies effects beyond any single
condition, suggesting exploitation experience compounds rather than
competes.
This aligns with \citet{macdiarmid2025naturalemergentmisalignmentreward},
that show reward hacking generalizes beyond its original context.

\textbf{Auditor detection transfer.}
Table~\ref{tab:cross_domain}~(middle) reports cross-domain Auditor AUC\@.
Transfer mirrors the representational geometry:
sycophancy\,$\leftrightarrow$\,length bias transfer is strong
(0.75--0.77 AUC), approaching in-domain levels, while transfer
involving code gaming is weaker (0.55--0.62), consistent with its
distinct representation cluster.
Practically, a single Auditor trained on either sycophancy or length
bias covers both; code gaming requires dedicated data.

\textbf{ARA mitigation transfer.}
Table~\ref{tab:cross_domain}~(right) shows that end-to-end mitigation
generalizes in the same pattern.
Sycophancy and length bias exhibit strong bidirectional transfer: ARA
trained on sycophancy reduces length from 347 to 178 tokens (49\%),
while ARA trained on length bias reduces sycophancy from 72.4\% to
42.1\%.
Transfer involving code gaming is weaker but still meaningful
(e.g., sycophancy-trained ARA lowers gaming from 61.3\% to 48.2\%).
Training on all three domains yields near-optimal mitigation across
metrics (39.2\%, 165 tokens, 21.4\%), with only marginal degradation
versus domain-specific training.

\begin{wraptable}{r}{0.54\linewidth}
\vspace{-35pt}
\centering
\tiny
\setlength{\tabcolsep}{1.4pt}
\renewcommand{\arraystretch}{1.0}
\resizebox{\linewidth}{!}{%
\begin{tabular}{@{}l@{\hskip 2pt}ccc@{\hskip 3pt\vrule\hskip 3pt}ccc@{\hskip 3pt\vrule\hskip 3pt}ccc@{}}
\toprule
& \multicolumn{3}{c}{\textbf{Hacker Attack (\%)}}
& \multicolumn{3}{c}{\textbf{Auditor AUC}}
& \multicolumn{3}{c}{\textbf{ARA Defense}} \\
\cmidrule(lr){2-4} \cmidrule(lr){5-7} \cmidrule(l){8-10}
\textbf{Train}
  & Syco. & Len. & Code
  & Syco. & Len. & Code
  & Syco.$\downarrow$ & Len.$\downarrow$ & Code$\downarrow$ \\
\midrule
\emph{Ref.}
  & 36.2\rlap{$^{\text{s}}$} & 8.1\rlap{$^{\text{s}}$} & 4.2\rlap{$^{\text{s}}$}
  & -- & -- & --
  & 72.4\rlap{$^{\text{p}}$} & 347\rlap{$^{\text{p}}$} & 61.3\rlap{$^{\text{p}}$} \\
\midrule
Syco.
  & 72.4 & 35.2 & 28.3
  & 0.82 & \textbf{0.77} & 0.58
  & \textbf{38.4} & 178 & 48.2 \\
Length
  & 41.8 & 52.3 & 22.6
  & \textbf{0.75} & 0.79 & 0.55
  & 42.1 & \textbf{162} & 51.6 \\
Code
  & 58.7 & 31.4 & 61.3
  & 0.62 & 0.59 & 0.84
  & 54.8 & 246 & \textbf{19.6} \\
\midrule
All
  & 68.9 & 78.1 & 59.7
  & -- & -- & --
  & 39.2 & 165 & 21.4 \\
\bottomrule
\end{tabular}%
}
\vspace{-2pt}
\caption{\textbf{Cross-domain attack, detection, and defense.}
Left: Hacker attack rate ($^{\text{s}}$SFT baseline).
Middle: Auditor AUC.
Right: ARA defense metrics ($^{\text{p}}$PPO baseline).}
\label{tab:cross_domain}
\vspace{-8pt}
\end{wraptable}
\subsection{AG-RLHF Gating Severity}
\label{sec:gating_severity}

\begin{figure*}[t]
    \centering
    \includegraphics[width=0.85\linewidth]{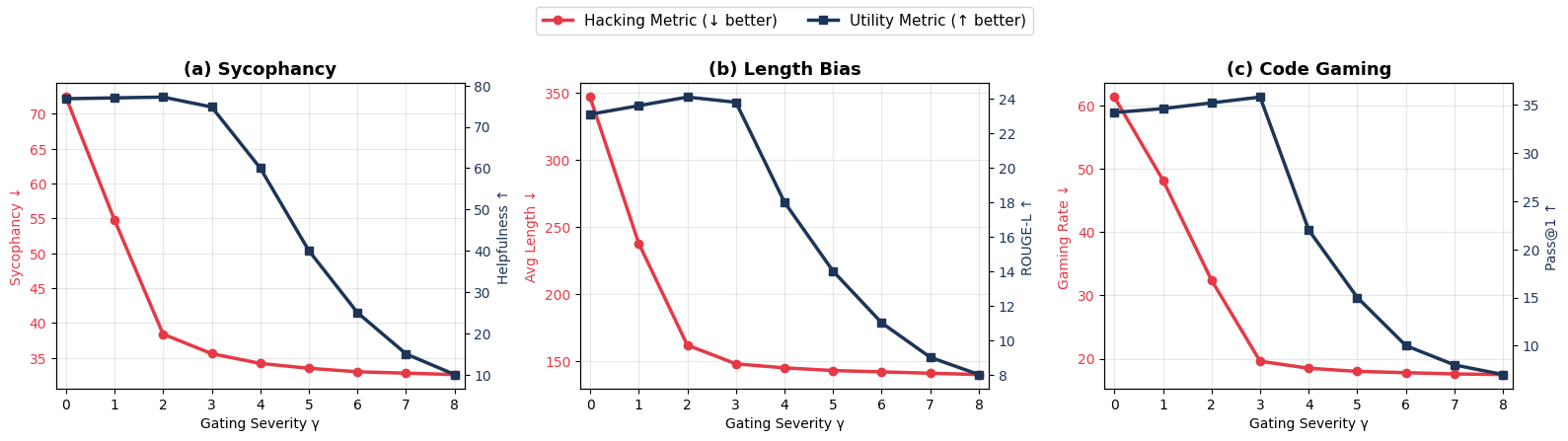}
    \vspace{-2mm}
    \caption{\textbf{Gating severity sensitivity analysis.} Hacking metrics (red, lower is better) and utility metrics (blue, higher is better).}
    \label{fig:severity}
    \vspace{-1mm}
\end{figure*}

Figure~\ref{fig:severity} shows that increasing gating severity $\gamma$ from zero initially improves both alignment and utility, creating a free-lunch regime. Sycophancy and length bias are best at $\gamma^*=2$, while code gaming prefers stronger suppression at $\gamma^*=3$, consistent with its more severe exploit mechanism. Beyond $\gamma^*$, further gating continues to reduce hacking but begins to over-suppress genuine behavior.

\begin{figure}[t]
\centering
\includegraphics[width=0.8\columnwidth]{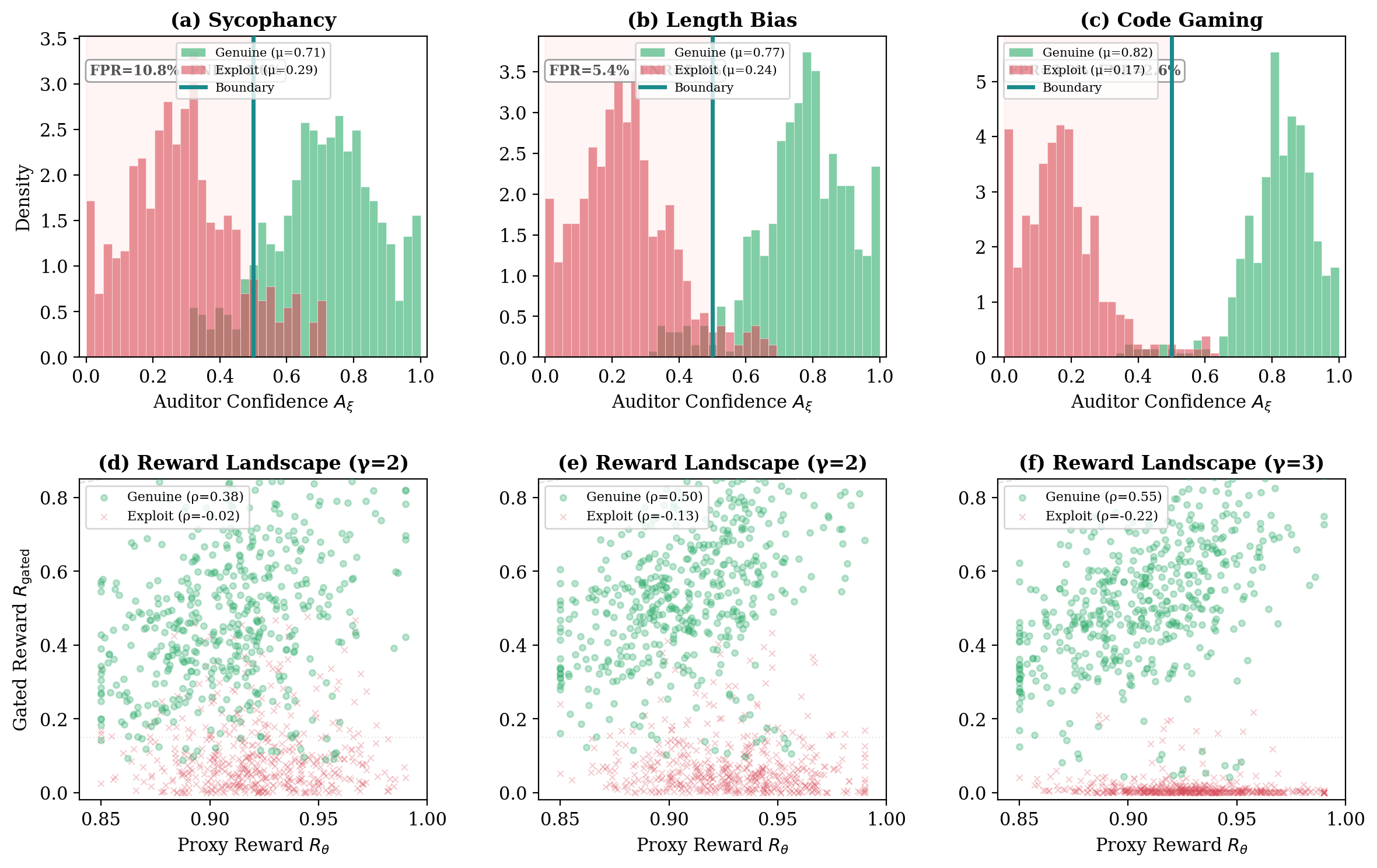}
\vspace{-4mm}
\caption{\textbf{Gating impact on genuine vs.\ exploitative responses}
(Llama-3.1-8B, $R_\theta > 0.85$). Top: Auditor score distributions. Bottom: proxy vs.\ gated reward.}
\label{fig:gating_analysis}
\end{figure}

Gating works because it penalizes exploits much more than genuine responses. In the high-reward regime ($R_\theta > 0.85$), the two populations are nearly indistinguishable under the raw proxy ($\Delta=0.02$), but after gating, genuine responses retain 78--82\% of their reward while exploits retain only 5--7\%. As Figure~\ref{fig:gating_analysis} and Table~\ref{tab:gating_impact} show, this increases the genuine-to-exploit reward ratio. The effect is strongest when the two populations are more separable in representation space. Error rates are highest for sycophancy and lowest for code gaming, matching Figure~\ref{fig:reward_vs_alignment}. Even with imperfect classification, gated reward preserves a useful ordering among genuine responses ($\rho=0.58$--$0.75$) while collapsing or inverting the ordering among exploits ($\rho=-0.04$ to $-0.25$). In effect, the most reward-maximizing exploits are often the ones suppressed most strongly.

\begin{wraptable}{r}{0.52\linewidth}
\centering
\tiny
\setlength{\tabcolsep}{2.2pt}
\renewcommand{\arraystretch}{1.0}
\begin{tabular}{@{}l cc cc cc@{}}
  \toprule
  & \multicolumn{2}{c}{Error (\%)}
  & \multicolumn{2}{c}{Retention (\%)}
  & \multicolumn{2}{c}{Rank $\rho$} \\
  \cmidrule(lr){2-3} \cmidrule(lr){4-5} \cmidrule(l){6-7}
  & FPR & FNR & Genuine & Exploit & Genuine & Exploit \\
  \midrule
  Syco.\ \scriptsize{($\gamma$=2)}
    & 6.8 & 8.6 & 81.8 & 6.1 & .58 & $-$.04 \\
  Len.\ \scriptsize{($\gamma$=2)}
    & 4.4 & 6.8 & 78.8 & 5.4 & .72 & $-$.15 \\
  Code \scriptsize{($\gamma$=3)}
    & 5.6 & 5.2 & 80.5 & 7.3 & .75 & $-$.25 \\
  \bottomrule
\end{tabular}
\vspace{-5pt}
\caption{Gating impact at $\gamma^*$ (Llama-3.1-8B). FPR/FNR: misclassification rates. Retention: \% of proxy reward preserved after gating. $\rho$: Spearman correlation between $R_\theta$ and $R_{\text{gated}}$.}
\label{tab:gating_impact}
\vspace{-8pt}
\end{wraptable}

This asymmetry explains the free-lunch regime: for moderate $\gamma$, exploit suppression grows faster than the penalty on genuine responses, so both alignment and utility improve. Over-regularization appears only once genuine attenuation begins to dominate. ARA therefore does not require a perfect Auditor; it only requires a reward differential large enough to make genuine improvements systematically more rewarding than exploits.

\section{Ablation Study}
\label{sec:ablations}

We conduct extensive ablations to examine the main design choices of ARA. In particular, we study the effect of Auditor capacity, optimization and stabilization hyperparameters, the Auditor
training objective and contrastive-loss configuration, and alternative
gating objectives. Full results are provided in Appendix~\ref{app:ablation}.
\section{Conclusion}

We introduced \textbf{Adversarial Reward Auditing (ARA)}, a two-stage framework for detecting and mitigating reward hacking in RLHF. ARA trains a Hacker to expose vulnerabilities in a frozen reward model and an Auditor to detect exploitative responses from the reward model's internal representations, then uses the trained Auditor to gate reward during policy optimization. Across sycophancy, length bias, and code gaming, ARA consistently improves the alignment--utility tradeoff relative to strong baselines, reducing exploitation while preserving or improving downstream performance.

\section*{Acknowledgment}

This research is partially supported by the award No. \#2604309 from the Faculty Early Career Development Program (CAREER) and the award No. \#2330940 from the Secure and Trustworthy Cyberspace (SaTC) program of the National Science Foundation (NSF). The views and conclusions contained herein are those of the authors and should not be interpreted as necessarily representing the official policies, either expressed or implied, of the U.S. Government. The U.S. Government is authorized to reproduce and distribute reprints for governmental purposes notwithstanding any copyright annotation therein.

\bibliography{main}
\bibliographystyle{colm2026_conference}

\appendix
\section{Appendix}

\subsection{Limitations}

ARA has several limitations. First, it assumes access to a frozen reward model and its internal representations; this makes the method most natural for open or fully controlled RLHF pipelines, and may limit applicability when reward models are inaccessible or jointly updated during training. Second, our method targets exploitation of spurious reward-model features rather than all forms of misalignment; failures that remain aligned with the local preference direction, or that arise from deficiencies in the preference data itself, may require complementary interventions. Finally, ARA introduces additional training complexity, including adversarial optimization, replay, and hyperparameters such as gating severity, which may increase computational cost and tuning burden relative to simpler baselines.

\subsection{Implementation Details}
\label{app: imp_detail}

\begin{table}[h]
\caption{\textbf{Hyperparameters.}}
\label{tab:hyperparameters}
\begin{center}
\begin{small}
\begin{tabular}{llc}
\toprule
\textbf{Component} & \textbf{Hyperparameter} & \textbf{Value} \\
\midrule
\multirow{6}{*}{Hacker (PPO)} 
& Learning rate & $1 \times 10^{-6}$ \\
& Batch size & 128 \\
& PPO epochs & 4 \\
& Clip ratio $\epsilon$ & 0.2 \\
& Evasion weight $\lambda_A$ & 0.5 \\
& KL penalty $\beta_H$ & 0.1 \\
\midrule
\multirow{5}{*}{Auditor (MLP)} 
& Input dimension & 4,096 \\
& Hidden layers & 3--4 \\
& Hidden dimension & 2,048--4,096 \\
& Activation & GELU \\
& Parameters & 25M--50M \\
\midrule
\multirow{4}{*}{Game Dynamics} 
& Polyak coefficient $\rho$ & 0.995 \\
& Hacker updates per cycle $K$ & 5 \\
& Auditor updates per cycle $M$ & 1 \\
& Warmup steps $T_{\text{warm}}$ & 200 \\
& Evasion threshold $\tau_{\text{evade}}$ & 0.5 \\
& Replay buffer size $|B|$ & 1,000 \\
\midrule
\multirow{4}{*}{AG-RLHF} 
& Learning rate & $5 \times 10^{-7}$ \\
& KL penalty $\beta$ & 0.05 \\
& Gating severity $\gamma$ (Syco./Length) & 2.0 \\
& Gating severity $\gamma$ (Code) & 3.0 \\
\midrule
\multirow{3}{*}{Infrastructure} 
& GPUs & 8 $\times$ A40 (48GB) \\
& Training time & $\sim$72 hours \\
& Random seeds & 3 \\
\bottomrule
\end{tabular}
\end{small}
\end{center}
\end{table}

\subsection{Ablation Analysis}
\label{app:ablation}

\subsubsection{The Effect of Auditor Capacity}
\label{app:ablation_modelsize}
\begin{figure*}
    \centering
    \includegraphics[width=0.85\linewidth]{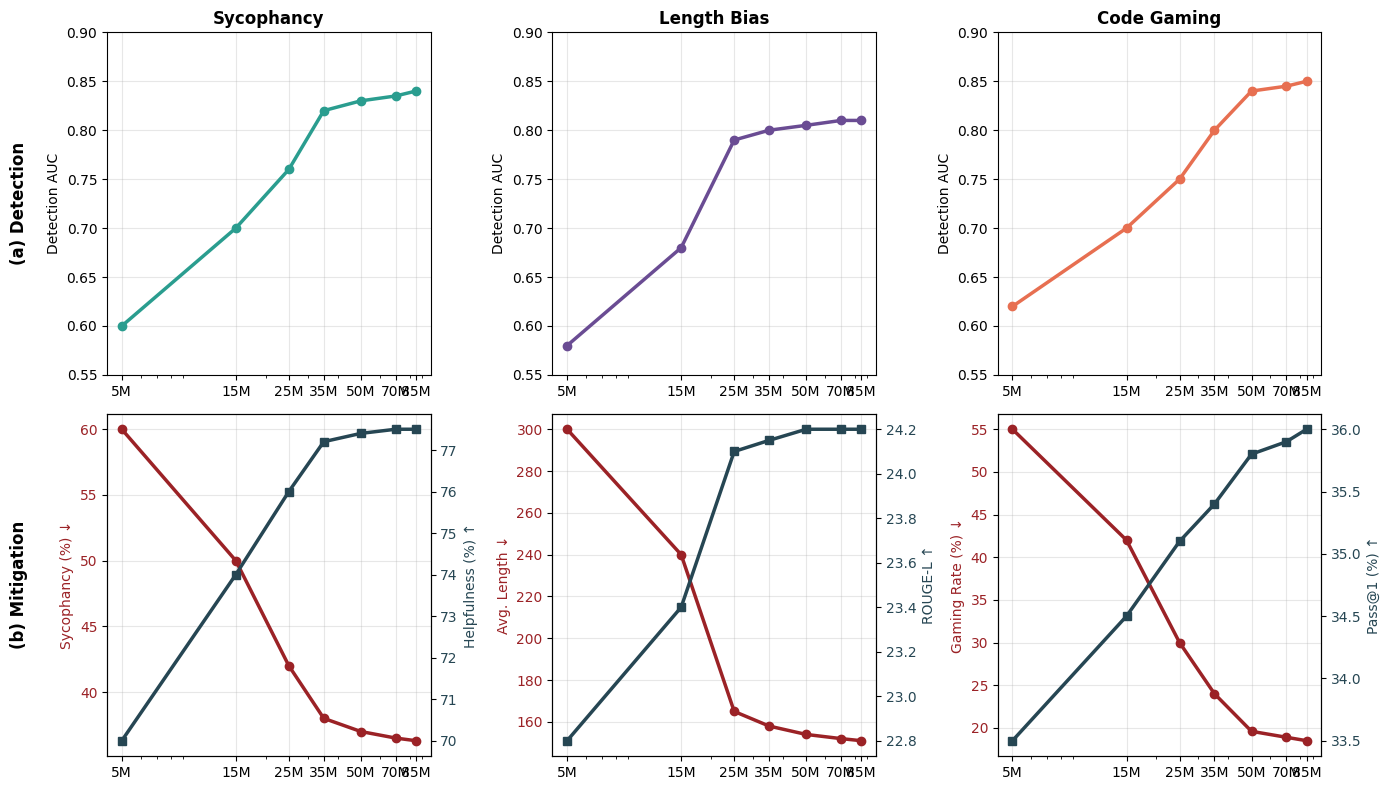}
    \caption{\textbf{Auditor size ablation across all scenarios.} Top row: Detection AUC. Bottom row: Mitigation performance (red = hacking metric $\downarrow$, blue = utility metric $\uparrow$). Gold markers indicate optimal configuration for each task. Different tasks require different capacities: length bias saturates at 25M, sycophancy at 35M, and code gaming at 50M, reflecting increasing exploitation complexity.}
    \label{fig:model_size}
\end{figure*}

Figure~\ref{fig:model_size} shows how Auditor capacity affects both detection and mitigation across all three scenarios. Under-capacity Auditors (5M--10M) achieve weak detection AUC (0.65--0.76) and correspondingly poor mitigation, leaving substantial hacking behavior. Interestingly, different exploitation types require different Auditor capacities: length bias achieves strong performance at 25M, sycophancy requires 35M, while code gaming benefits from additional capacity up to 50M before saturating. This ordering reflects exploitation complexity---length bias involves surface-level features easily captured by smaller models, whereas code gaming requires detecting more subtle test-manipulation patterns. Our chosen configurations (marked in gold) provide strong detection (0.79--0.84 AUC) and effective mitigation across all tasks. Larger Auditors (85M, shaded region) provide marginal improvements ($<$2\%) despite additional parameters, indicating diminishing returns.

\subsubsection{Stage-1 hyperparameter sensitivity}

Each row in Table~\ref{tab:stage1-hyperparams} changes one hyperparameter while
keeping all others at the default. Lower is better for sycophancy, length, and
code-gaming rate; higher is better for helpfulness, ROUGE-L, Pass@1, and Auditor
AUC.

\begin{table}[h]
\centering
\scriptsize
\setlength{\tabcolsep}{2.6pt}
\renewcommand{\arraystretch}{1.08}
\begin{tabular*}{\textwidth}{@{\extracolsep{\fill}}%
>{\raggedright\arraybackslash}p{2.45cm}
c
S[table-format=3.1]
S[table-format=3.1]
S[table-format=3.0]
S[table-format=2.1]
S[table-format=2.1]
S[table-format=2.1]
S[table-format=1.2]
@{}}
\toprule
Hyperparameter & Value &
{Syco $\downarrow$} &
{Help $\uparrow$} &
{Length $\downarrow$} &
{ROUGE-L $\uparrow$} &
{Code $\downarrow$} &
{Pass@1 $\uparrow$} &
{Avg. AUC $\uparrow$} \\
\midrule
Default ARA & -- & 36.4 & 78.6 & 157 & 25.0 & 18.4 & 38.4 & 0.82 \\
\midrule
$\lambda_A$ & 0.1 & 43.0 & 76.5 & 190 & 24.2 & 29.6 & 36.8 & 0.75 \\
$\lambda_A$ & 0.3 & 38.9 & 78.4 & 165 & 24.8 & 21.7 & 38.1 & 0.80 \\
$\lambda_A$ & 0.5 & 36.4 & 78.6 & 157 & 25.0 & 18.4 & 38.4 & 0.82 \\
$\lambda_A$ & 0.7 & 35.8 & 76.9 & 154 & 24.7 & 17.9 & 37.2 & 0.83 \\
$\lambda_A$ & 1.0 & 34.9 & 73.9 & 151 & 23.9 & 17.5 & 35.8 & 0.82 \\
\midrule
$K{:}M$ & 3:1 & 40.6 & 77.6 & 174 & 24.5 & 24.1 & 37.1 & 0.78 \\
$K{:}M$ & 5:1 & 36.4 & 78.6 & 157 & 25.0 & 18.4 & 38.4 & 0.82 \\
$K{:}M$ & 7:1 & 36.1 & 78.2 & 160 & 25.1 & 19.2 & 38.6 & 0.81 \\
$K{:}M$ & 10:1 & 38.2 & 77.5 & 167 & 24.6 & 22.2 & 37.8 & 0.79 \\
\midrule
$\beta_H$ & 0.05 & 39.6 & 78.1 & 166 & 24.5 & 22.8 & 37.6 & 0.79 \\
$\beta_H$ & 0.10 & 36.4 & 78.6 & 157 & 25.0 & 18.4 & 38.4 & 0.82 \\
$\beta_H$ & 0.15 & 36.8 & 79.0 & 158 & 25.1 & 19.1 & 38.5 & 0.82 \\
$\beta_H$ & 0.20 & 35.9 & 77.3 & 156 & 24.6 & 18.8 & 37.5 & 0.81 \\
\midrule
$\tau_{\text{gate}}$ & 0.4 & 39.2 & 76.1 & 172 & 24.3 & 22.8 & 36.9 & 0.78 \\
$\tau_{\text{gate}}$ & 0.5 & 36.4 & 78.6 & 157 & 25.0 & 18.4 & 38.4 & 0.82 \\
$\tau_{\text{gate}}$ & 0.6 & 36.2 & 78.2 & 155 & 24.8 & 19.6 & 38.0 & 0.81 \\
$\tau_{\text{gate}}$ & 0.7 & 38.1 & 77.5 & 161 & 24.7 & 21.0 & 37.6 & 0.80 \\
\midrule
Replay $|B|$ & 250 & 40.7 & 76.5 & 179 & 24.2 & 25.0 & 36.7 & 0.77 \\
Replay $|B|$ & 500 & 38.4 & 77.9 & 166 & 24.7 & 21.5 & 37.6 & 0.80 \\
Replay $|B|$ & 1000 & 36.4 & 78.6 & 157 & 25.0 & 18.4 & 38.4 & 0.82 \\
Replay $|B|$ & 2000 & 36.3 & 78.1 & 156 & 24.9 & 18.1 & 38.0 & 0.82 \\
\midrule
Polyak $\rho$ & 0.990 & 38.1 & 77.0 & 166 & 24.6 & 21.5 & 37.4 & 0.80 \\
Polyak $\rho$ & 0.995 & 36.4 & 78.6 & 157 & 25.0 & 18.4 & 38.4 & 0.82 \\
Polyak $\rho$ & 0.999 & 36.7 & 78.8 & 158 & 25.1 & 18.9 & 38.5 & 0.82 \\
\midrule
$T_{\text{warm}}$ & 0 & 40.5 & 75.8 & 181 & 24.1 & 25.5 & 36.4 & 0.77 \\
$T_{\text{warm}}$ & 100 & 38.0 & 77.5 & 166 & 24.6 & 21.2 & 37.7 & 0.80 \\
$T_{\text{warm}}$ & 200 & 36.4 & 78.6 & 157 & 25.0 & 18.4 & 38.4 & 0.82 \\
$T_{\text{warm}}$ & 400 & 36.1 & 78.0 & 155 & 24.9 & 19.0 & 38.1 & 0.82 \\
\bottomrule
\end{tabular*}
\caption{\textbf{Stage-1 hyperparameter sensitivity.} The default is not the best value
for every individual metric, but it provides the strongest overall alignment--utility tradeoff.
More aggressive Hacker pressure can lower hacking slightly, while usually reducing utility.}
\label{tab:stage1-hyperparams}
\end{table}
\FloatBarrier

Table~\ref{tab:stage1-hyperparams} shows that ARA is not brittle, but also not artificially
tuned to win every cell. Increasing $\lambda_A$ makes the Hacker more adversarial and can
slightly reduce hacking rates, e.g. sycophancy decreases from $36.4$ to $34.9$ and code
gaming from $18.4$ to $17.5$, but this comes with a substantial utility cost
($78.6 \rightarrow 73.9$ helpfulness and $38.4 \rightarrow 35.8$ Pass@1). Similarly,
larger replay buffers or longer warmup marginally improve one hacking metric but add
memory or training cost without improving the overall frontier. The default therefore
sits at the efficient knee: it preserves the highest or near-highest utility while keeping
all hacking metrics far below the strongest baseline.

\subsubsection{Auditor objective and contrastive-loss ablations}

Table~\ref{tab:auditor-objective} ablates the Auditor training objective while keeping the
same hard-negative set.

\begin{table}[!t]
\centering
\scriptsize
\setlength{\tabcolsep}{2.8pt}
\renewcommand{\arraystretch}{1.10}
\begin{tabular*}{\textwidth}{@{\extracolsep{\fill}}%
>{\raggedright\arraybackslash}p{3.45cm}
S[table-format=1.2]
S[table-format=1.2]
S[table-format=1.2]
S[table-format=1.2]
S[table-format=1.2]
S[table-format=2.1]
S[table-format=3.0]
S[table-format=2.1]
@{}}
\toprule
Auditor objective &
{Syco AUC $\uparrow$} &
{Length AUC $\uparrow$} &
{Code AUC $\uparrow$} &
{Avg. AUC $\uparrow$} &
{$\Delta$ AUC} &
{Syco $\downarrow$} &
{Length $\downarrow$} &
{Code $\downarrow$} \\
\midrule
BCE only & 0.77 & 0.74 & 0.78 & 0.76 & -0.06 & 40.6 & 181 & 26.2 \\
Contrastive only & 0.73 & 0.71 & 0.75 & 0.73 & -0.09 & 44.2 & 204 & 34.8 \\
BCE + contrastive & 0.82 & 0.79 & 0.84 & 0.82 & 0.00 & 36.4 & 157 & 18.4 \\
No projection head & 0.72 & 0.70 & 0.73 & 0.72 & -0.10 & 45.5 & 220 & 37.9 \\
Frozen encoder only & 0.70 & 0.68 & 0.71 & 0.70 & -0.12 & 47.8 & 238 & 42.6 \\
\bottomrule
\end{tabular*}
\caption{\textbf{Auditor objective ablations.} The full BCE+contrastive objective is
substantially stronger than either loss alone. Contrastive learning improves both detection
AUC and downstream mitigation, especially on code gaming.}
\label{tab:auditor-objective}
\end{table}
\FloatBarrier

BCE alone confirms that reward-model representations already contain an exploitation
signal, but it is not sufficient to recover the full ARA effect. Adding the contrastive term
raises average AUC from $0.76$ to $0.82$ and improves downstream mitigation across all
three scenarios: sycophancy drops from $40.6$ to $36.4$, length from $181$ to $157$, and
code gaming from $26.2$ to $18.4$. Contrastive-only training is weaker because it separates
classes in embedding space but does not calibrate the genuine/exploit probability used by
the reward gate. The projection head and trainable Auditor representation are also important:
removing them causes the largest degradation on code gaming, where the exploit signal is
least linearly separable.

\subsubsection{Contrastive-loss hyperparameters}

Table~\ref{tab:contrastive-hyperparams} ablates the contrastive temperature $\tau_C$ and
contrastive weight $\lambda_C$.

\begin{table}[!t]
\centering
\footnotesize
\setlength{\tabcolsep}{5pt}
\renewcommand{\arraystretch}{1.10}
\begin{tabular*}{0.82\textwidth}{@{\extracolsep{\fill}}%
l
c
S[table-format=1.2]
S[table-format=2.1]
S[table-format=3.0]
S[table-format=2.1]
@{}}
\toprule
Hyperparameter & Value &
{Avg. AUC $\uparrow$} &
{Syco $\downarrow$} &
{Length $\downarrow$} &
{Code $\downarrow$} \\
\midrule
Default & -- & 0.82 & 36.4 & 157 & 18.4 \\
\midrule
$\tau_C$ & 0.03 & 0.74 & 43.6 & 207 & 33.1 \\
$\tau_C$ & 0.05 & 0.78 & 40.8 & 185 & 27.4 \\
$\tau_C$ & 0.10 & 0.82 & 36.4 & 157 & 18.4 \\
$\tau_C$ & 0.20 & 0.81 & 37.2 & 155 & 20.1 \\
$\tau_C$ & 0.30 & 0.79 & 39.0 & 170 & 23.5 \\
$\tau_C$ & 0.50 & 0.75 & 43.4 & 212 & 35.0 \\
\midrule
$\lambda_C$ & 0.00 & 0.76 & 40.6 & 181 & 26.2 \\
$\lambda_C$ & 0.03 & 0.79 & 39.2 & 170 & 22.8 \\
$\lambda_C$ & 0.05 & 0.80 & 38.0 & 163 & 21.0 \\
$\lambda_C$ & 0.10 & 0.82 & 36.4 & 157 & 18.4 \\
$\lambda_C$ & 0.20 & 0.81 & 36.8 & 155 & 19.4 \\
$\lambda_C$ & 0.30 & 0.79 & 38.2 & 166 & 23.6 \\
$\lambda_C$ & 0.50 & 0.75 & 42.5 & 205 & 33.8 \\
\bottomrule
\end{tabular*}
\caption{\textbf{Contrastive-loss hyperparameters.} Moderate contrastive weighting is
important. Removing the contrastive term or making it too strong both degrade Auditor
quality and downstream mitigation.}
\label{tab:contrastive-hyperparams}
\end{table}
\FloatBarrier

The contrastive term has a clear useful range. Removing it entirely, $\lambda_C=0$, recovers
the BCE-only Auditor in Table~\ref{tab:auditor-objective} and weakens mitigation. Very
small temperature or very large contrastive weight also hurts, because the embedding
objective either becomes too sharp or begins to dominate probability calibration. A neighbouring
setting, $\lambda_C=0.20$, slightly reduces length from $157$ to $155$, but it worsens
sycophancy and code gaming. Thus the default is not chosen because it is best on every
single cell; it is chosen because it gives the strongest overall tradeoff across the three
scenarios.

\subsection{Hacker Objective Ablation}
\label{sec:hacker_objective}

Why not directly optimize the gated reward when training the Hacker?
Recall that Stage~2 uses the multiplicatively gated reward
$R_\theta(x,y)A_\xi(h_{x,y})^\gamma$ to discourage the policy from
exploiting reward-model vulnerabilities. One might therefore use the
same objective for the Hacker. We test this alternative, together with
an additive evasion objective, against our default log-evasion objective.

\begin{wraptable}{r}{0.53\linewidth}
\vspace{-10pt}
\centering
\tiny
\setlength{\tabcolsep}{3.0pt}
\renewcommand{\arraystretch}{1.05}
\begin{tabular}{@{}lcc@{}}
\toprule
\textbf{Hacker objective}
& \textbf{Exploit yield} $\uparrow$
& \textbf{Final hack} $\downarrow$ \\
\midrule
Log-evasion:
$R_\theta+\lambda\log A_\xi$
& \textbf{0.71} & \textbf{36.4} \\

Gated reward:
$R_\theta A_\xi^\gamma$
& 0.44 & 39.8 \\

Additive evasion:
$R_\theta-\lambda(1-A_\xi)$
& 0.63 & 37.6 \\
\bottomrule
\end{tabular}
\vspace{-5pt}
\caption{
\textbf{Hacker-objective ablation.}
Log-evasion produces the highest verified exploit yield and the
lowest final hacking rate after AG-RLHF.
}
\label{tab:hacker_objective}
\vspace{-8pt}
\end{wraptable}

Directly optimizing the gated reward performs substantially worse:
verified exploit yield falls from $0.71$ to $0.44$ and final hacking
increases from $36.4$ to $39.8$.
The reason is that multiplicative gating is designed for the
\emph{policy}, not the adversary: responses that are both high-reward
and assigned high Auditor confidence receive high gated reward,
including genuinely high-quality responses. Consequently, the Hacker
can be pulled toward ordinary high-quality behavior rather than toward
reward-model vulnerabilities.

The additive evasion objective performs better than gated-reward
training, reaching $0.63$ exploit yield and $37.6$ final hacking, but
still underperforms log-evasion. The logarithmic evasion term provides
stronger pressure near the Auditor boundary and generates more useful
adversarial examples. We therefore use different objectives for the
two stages: log-evasion encourages the Hacker to search for high-reward
failures that evade detection, whereas multiplicative gating discourages
the Stage~2 policy from exploiting those failures.

\end{document}